\documentclass[10pt,twocolumn]{article}
\usepackage[top=2cm, bottom=2cm, left=1.85cm, right=1.85cm]{geometry}
\usepackage{times}  %
\usepackage[sort&compress,numbers]{natbib}  %
\usepackage[hyphens]{url}
\usepackage{graphicx}  %
\usepackage[keeplastbox]{flushend}
\usepackage[hyphens]{url}  %
\usepackage{graphicx} %
\usepackage{tikzsymbols}

\usepackage{algorithm}
\usepackage{algorithmic}

\let\oldbibliography\thebibliography
\renewcommand{\thebibliography}[1]{%
  \oldbibliography{#1}%
  \setlength{\itemsep}{2pt}%
}

\usepackage[hang,flushmargin]{footmisc}

\usepackage[compact]{titlesec}
\titlespacing*{\section}{0pt}{*3}{3pt}
\titlespacing*{\subsection}{0pt}{*2}{2pt}
\usepackage{xspace}

\makeatletter
\def\url@leostyle{%
  \@ifundefined{selectfont}{\def\UrlFont{}}%
  {\def\UrlFont{}}%
}
\makeatother
\usepackage[hyphenbreaks]{breakurl}

\usepackage[bookmarks=true, bookmarksnumbered=true, colorlinks=true, linkcolor=linkcol, citecolor=citecol, urlcolor=urlcol, hypertexnames=true]{hyperref}

\definecolor{darkgreen}{RGB}{0, 100, 0}
\definecolor{linkcol}{rgb}{0.3,0,0}
\definecolor{citecol}{rgb}{0.3,0,0}
\definecolor{urlcol}{rgb}{0.3,0,0}

\makeatletter
\def\url@leostyle{%
  \@ifundefined{selectfont}{\def\UrlFont{\small}}%
  {\def\UrlFont{}}%
}
\makeatother
\usepackage[small,bf]{caption}
\usepackage{bbold}
\usepackage{amsmath}
\usepackage{color}
\usepackage{graphicx}  %
	\setkeys{Gin}{width=\textwidth, height=\textheight, keepaspectratio}
\usepackage[skip=0pt]{subcaption}

\newif\ifcomment
\commenttrue
\commentfalse

\ifcomment
	\newcommand{\edc}[1]{\textbf{\em\color{red}EDC: #1}}
	\newcommand{\bo}[1]{\textbf{\em\color{green}BO: #1}}
	\newcommand{\gvg}[1]{\textbf{\em\color{blue}GG: #1}}
\else
		\newcommand\edc[1]{}
		\newcommand\bo[1]{}
		\newcommand\gvg[1]{}
\fi
\newcommand{\descr}[1]{\smallskip\noindent\textbf{#1}}

\begin{document}

\title{DP-SGD vs PATE: Which Has Less Disparate Impact on GANs?}

\author{
 		Georgi Ganev\textsuperscript{\rm 1,}\textsuperscript{\rm 2}\\[0.5ex]
\normalsize 		\textsuperscript{\rm 1}UCL\;\;
 		\textsuperscript{\rm 2}Hazy\\
\normalsize 		georgi.ganev.16@ucl.ac.uk
 }

\date{}

\maketitle

\begin{abstract}
Generative Adversarial Networks (GANs) are among the most popular approaches to generate synthetic data, especially images, for data sharing purposes.
Given the vital importance of preserving the privacy of the individual data points in the original data, GANs are trained utilizing frameworks with robust privacy guarantees such as Differential Privacy~(DP).
However, these approaches remain widely unstudied beyond single performance metrics when presented with imbalanced datasets.
To this end, we systematically compare GANs trained with the two best-known DP frameworks for deep learning, DP-SGD, and PATE, in different data imbalance settings from two perspectives -- the size of the classes in the generated synthetic data and their classification performance.

Our analyses show that applying PATE, similarly to DP-SGD, has a disparate effect on the under/over-represented classes but in a much milder magnitude making it more robust.
Interestingly, our experiments consistently show that for PATE, unlike DP-SGD, the privacy-utility trade-off is not monotonically decreasing but is much smoother and inverted U-shaped, meaning that adding a small degree of privacy actually helps generalization.
However, we have also identified some settings (e.g., large imbalance) where PATE-GAN completely fails to learn some subparts of the training data.
\end{abstract}

\section{Introduction}

Generative machine learning models, and in particular Generative Adversarial Networks (GANs)~\cite{goodfellow2014generative}, have received increasing attention from both researchers~\cite{szpruch19synthetic, schaar20synthetic} and government organizations~\cite{benedetto2018creation, nist2018the, nist2018differential, nhs21ae} as a promising solution to the individual-level data sharing problem.
The underlying idea is to train generative models to learn the distribution of the (real) data, generate new high-quality (synthetic) samples from the trained model, and release synthetic, rather than real, data.

However, recent research has shown that generative models, including GANs, may leak sensitive information about the training samples through overfitting and memorization~\cite{carlini2019secret, webster2019detecting, meehan2020non} as well as susceptibility to privacy attacks such as membership inference attacks~\cite{hayes2019logan, chen2020gan, stadler2020synthetic}.
The state-of-the-art approach to protect against such vulnerabilities is training the models to satisfy Differential Privacy (DP)~\cite{dwork2006calibrating, dwork2014algorithmic}.
DP mechanisms protect against attempts to infer the inclusion of any record in the training data by bounding their individual contribution, usually through perturbation.
In this paper, we will focus on the two most widely used DP techniques for training deep learning models -- DP-SGD~\cite{abadi2016deep} and PATE~\cite{papernot2016semi, papernot2018scalable}.

Even though DP mechanisms guarantee rigorous privacy protection, they degrade the performance of the model.
Furthermore, this accuracy drop is likely to be disparate, affecting the underrepresented subpopulations of the data disproportionately more.
For example, in the case of deep learning classifiers~\cite{bagdasaryan2019differential, farrand2020neither, suriyakumar2021chasing} empirically illustrate the disparate degradation caused by DP-SGD.
However, comparisons between DP-SGD and PATE are still relatively unstudied in this light, with Uniyal et al.~\cite{uniyal2021dp} doing so for classifiers, and more recently, Ganev et al.~\cite{ganev2021robin} demonstrating the said effects in generative models trained on imbalanced tabular data.
To fill this gap, we set out to examine and compare two GAN models trained with DP guarantees (DP-WGAN and PATE-GAN) on imbalanced image data (MNIST) in several imbalance settings.

\descr{Research Question.}
Does applying DP-SGD and PATE to GANs lead to similar disparate effects when trained on imbalanced data, or more specifically, on the minority and majority classes in terms of size and accuracy of the resulting synthetic data?

\descr{Main Findings.}
Our experiments could be summarized to:
\begin{itemize}
	\item Overall, both models exhibit disparity in terms of size and accuracy, but the effects are much smaller for PATE-GAN.
	Furthermore, PATE-GAN offers a much better privacy-utility trade-off and performs better even for tight privacy budgets ($0.5$).
	We believe this is due to the teacher-discriminators setup.
	\item In terms of size, the two models behave in opposite directions with increased privacy -- DP-WGAN ``evens'' the classes while PATE-GAN increases the imbalance.
	\item In the presence of a single highly imbalanced class (``8'' reduced to 10\% its original size), PATE-GAN fails to learn the whole subpopulation.
	This is not the case for DP-WGAN.
	\item Applying some degree of privacy actually serves as regularization to PATE-GAN and helps the performance up to a point, unlike DP-WGAN, for which any privacy protection deteriorates the utility.
\end{itemize}

\section{Preliminaries}
In this section, we present some background on DP, GANs, and the two generative models used in our experiments.

\subsection{Differential Privacy (DP)}

A randomized algorithm $\mathcal{A}$ satisfies ($\epsilon, \delta$)-DP if and only if, for any two adjacent datasets $D_{1}$ and $D_{2}$ (differing in a single record), and all possible outputs $S$ of $\mathcal{A}$, the following holds~\cite{dwork2006calibrating, dwork2014algorithmic}:
\begin{equation*}
P[{\mathcal{A}}(D_{1})\in S]\leq \exp \left(\epsilon \right)\cdot P[{\mathcal{A}}(D_{2})\in S] + \delta \end{equation*}
Put simply, one cannot distinguish whether any individual's data was part of the input dataset from observing the algorithm's output.
The privacy budget $\epsilon$ denotes the level of indistinguishability while $\delta$ is a probability of privacy failure.
We focus on the two most widely used DP techniques for deep learning -- DP-SGD~\cite{abadi2016deep} and PATE~\cite{papernot2016semi, papernot2018scalable}.

\subsection{Generative Adversarial Networks (GANs)}

A GAN is a deep learning model consisting of two neural networks, a generator, and a discriminator.
They ``compete'' against each other in a min-max ``game'', the former produces synthetic data while the latter distinguishes real from generated samples until they reach equilibrium.

\descr{DP-WGAN.}
DP-WGAN~\cite{alzantot2019differential} utilizes DP-SGD in place of the standard SGD during training.
DP-SGD guarantees privacy by bounding the individual gradients (using clipping and perturbation) of the discriminator and relying on the moments accountant method to track the overall privacy budget.

Furthermore, the model uses the WGAN architecture~\cite{arjovsky2017wasserstein} to improve stability during training.

\descr{PATE-GAN.}
PATE-GAN~\cite{jordon2018pate} modifies the PATE framework for training GANs.
The model replaces the standard single discriminator with $k$ teacher-discriminators, trained on disjoint partitions of the real data.
In turn, the standard student model is replaced by a student-discriminator trained on noisy (real/synthetic) labels predicted by the teachers.
The noisy aggregation is where DP guarantees the privacy protection.
Also, the proposed model eliminates the need for publicly available data.

Even though both DP-WGAN and PATE-GAN were initially proposed for tabular data given the remarkable success of GANs on images, we decided not to adjust their architectures.

\section{Experimental Evaluation}
\label{sec:evaluation}

In this section, we explain our evaluation methodology and
discuss the experimental findings.

\subsection{Evaluation Methodology}
\begin{figure}[t!]
	\centering
	\begin{subfigure}{0.99\linewidth}
		\includegraphics{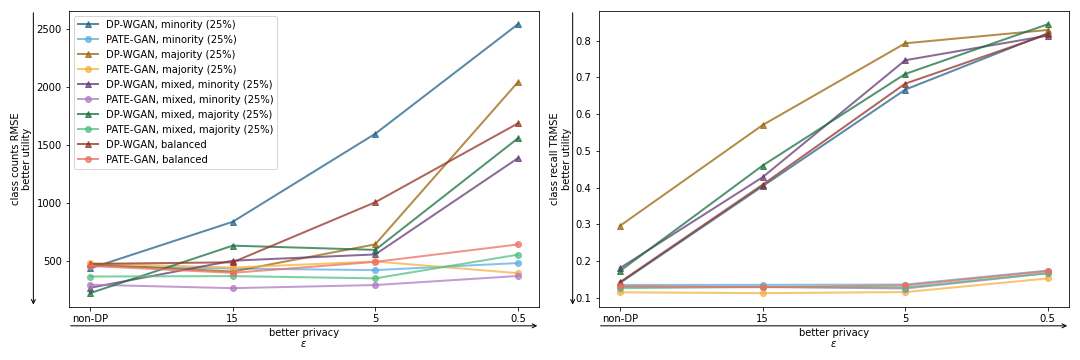}
	\end{subfigure}
	\caption{Class size RMSE (left) and recall TRMSE (right) vs $\epsilon$ trade-off of synthetic data generated by DP-WGAN and PATE-GAN in all settings.}
	\label{fig:mnist_trade_off}
\end{figure}

We run all of our experiments on MNIST~\cite{lecun2010mnist} and imbalance the class ``8''
to maintain consistency with~\cite{papernot2016semi, bagdasaryan2019differential, uniyal2021dp}.
First, since the classes are slightly imbalanced, we get rid of all images per class exceeding 5,000.
Then, we imbalance the dataset, making ``8'' the minority/majority class in one of the following three settings\footnote{we also experiment with balanced settings in App.~\ref{sec:balanced}}:
\begin{enumerate}
	\item {\em Minority} -- undersample ``8'' to 10\%/25\% its original size while keeping the other classes balanced.
	\item {\em Majority} -- keep ``8'' untouched but undersample all other classes to 10\%/25\% their sizes.
	\item {\em Mixed} -- turn ``8'' into minority/majority class by making it 25\% of the largest/smallest class and randomly imbalance all other classes in a uniformly decreasing manner.
\end{enumerate}
We train 5 DP-WGAN and PATE-GAN models for each setting, generate 5 synthetic datasets with a size equal to the input data (per trained model) and report mean and standard deviations.
We experiment with privacy budgets ($\epsilon$) of 0.5, 5, 15, and infinity (``non-DP'').
We measure the class distributions in the resulting synthetic datasets as well as class recall from classifiers (logistic regression similar to~\cite{ganev2021robin}) trained on the real/synthetic data and tested on put-aside test data.
We also report RMSE for sizes and truncated\footnote{we do not want to penalize the score if the recall achieved on the synthetic data is better than on the real} RMSE (TRMSE) for recall weighted by the real sizes in App.~\ref{sec:tables}.
A summary of the privacy-utility trade-off for both models in all settings is plotted in Fig.~\ref{fig:mnist_trade_off}.

Last, for PATE-GAN in all settings, we also experiment with a different number of teachers -- 1, 10, 50, and 100.

We use the same hyperparameters as the original implementations and set $\delta=10^{-5}$ for all experiments.

\subsection{Minority Class Results}
\label{ssec:min_results}

\begin{figure*}[t!]
	\centering
	\begin{subfigure}{0.495\linewidth}
		\includegraphics{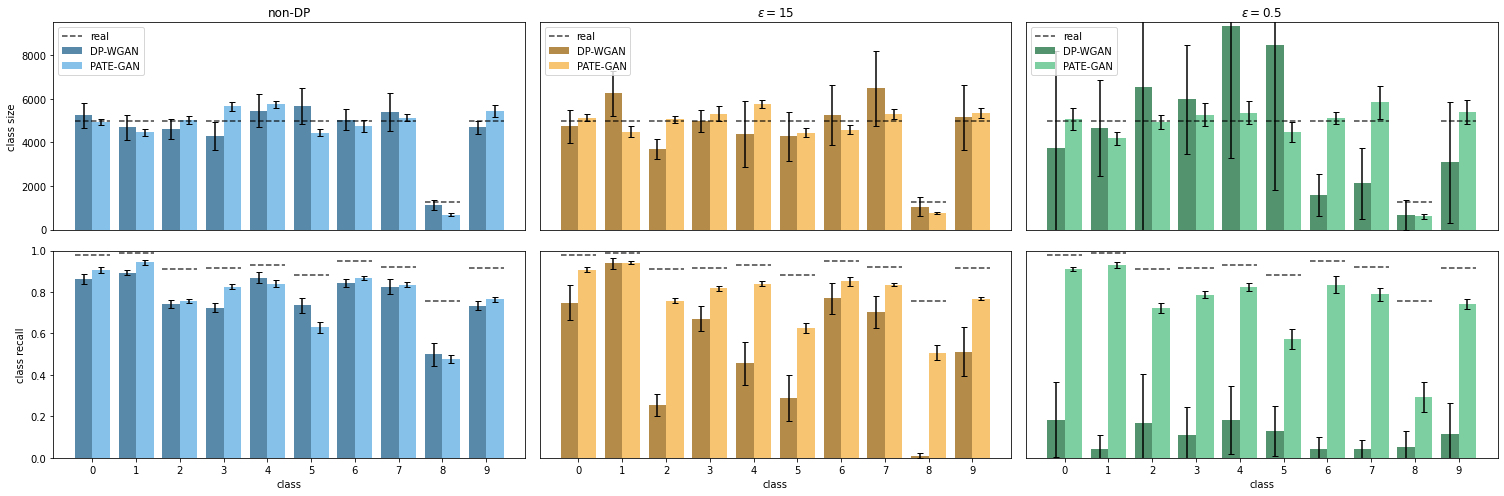}
		\caption{Minority (25\%)}
		\label{fig:mnist_min_0.25}
	\end{subfigure}
	\begin{subfigure}{0.495\linewidth}
		\includegraphics{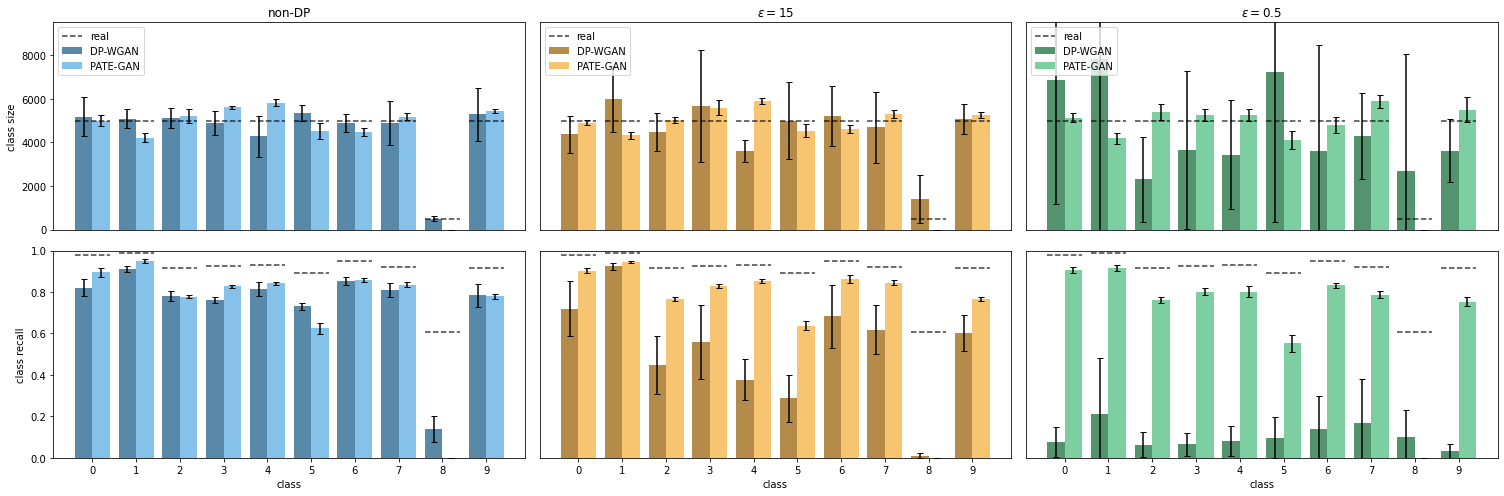}
		\caption{Minority (10\%)}
		\label{fig:mnist_min_0.1}
	\end{subfigure}
	\caption{Class size (top) and recall (bottom) of synthetic data generated by DP-WGAN and PATE-GAN trained with different privacy budgets $\epsilon$ on imbalanced MNIST, where class ``8'' is decreased to 25\% and 10\% of its original size.}
	\label{fig:mnist_min}
\end{figure*}

\begin{figure*}[t!]
	\centering
	\begin{subfigure}{0.495\linewidth}
		\includegraphics{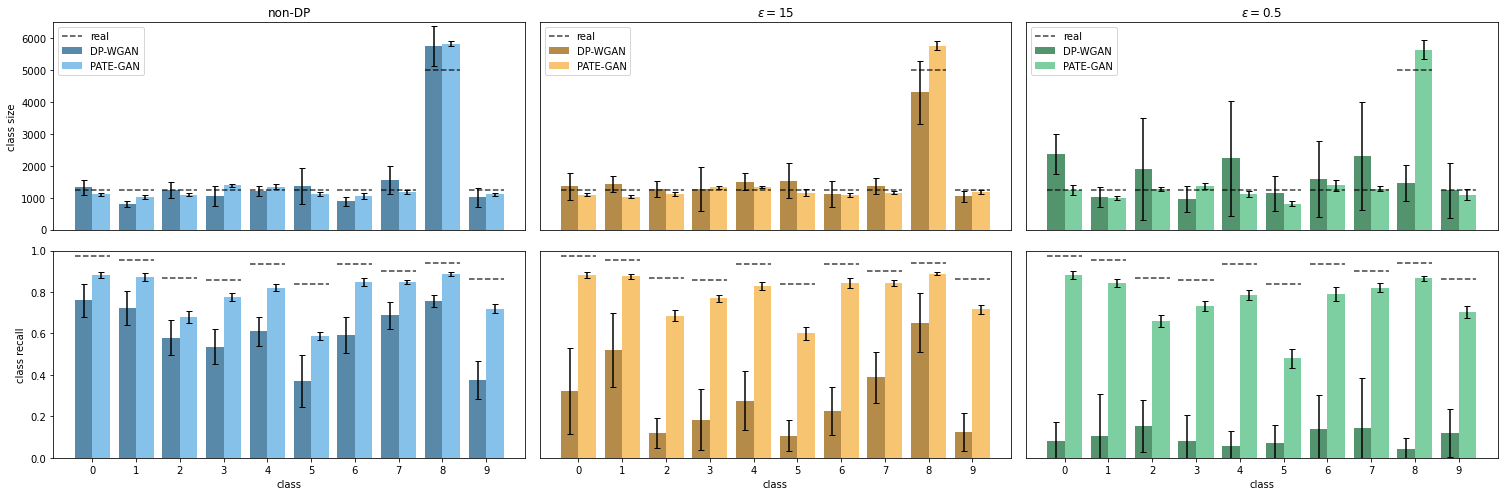}
		\caption{Majority (25\%)}
		\label{fig:mnist_max_0.25}
	\end{subfigure}
	\begin{subfigure}{0.495\linewidth}
		\includegraphics{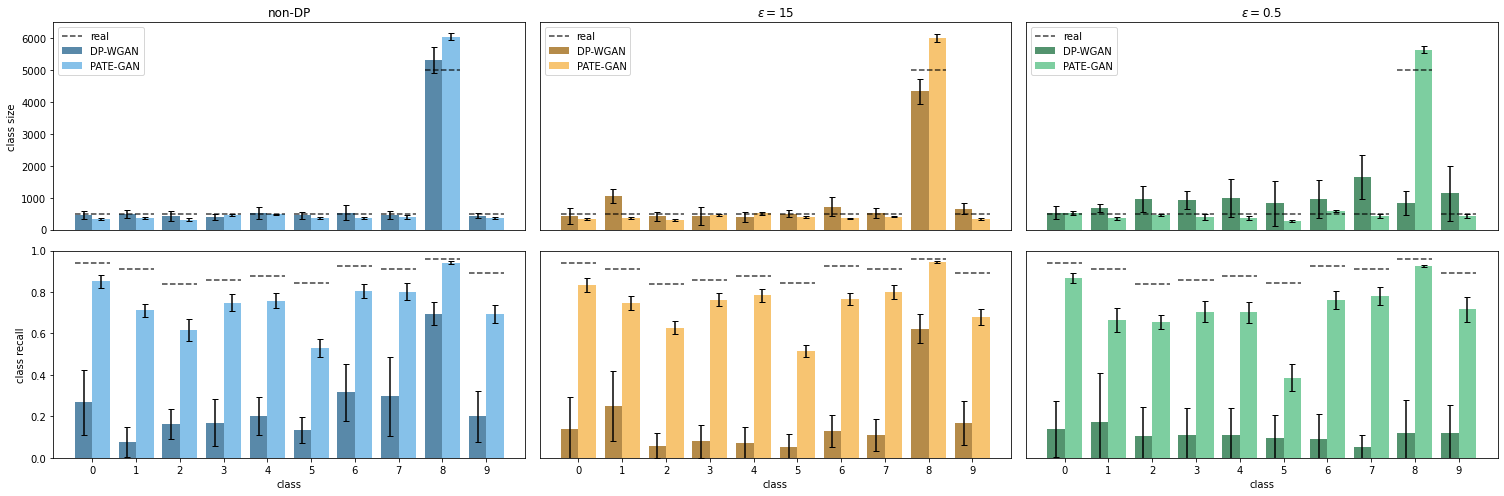}
		\caption{Majority (10\%)}
		\label{fig:mnist_max_0.1}
	\end{subfigure}
	\caption{Class size (top) and recall (bottom) of synthetic data generated by DP-WGAN and PATE-GAN trained with different privacy budgets $\epsilon$ on imbalanced MNIST, where all classes apart from ``8'' are decreased to 25\% and 10\% of their original size.}
	\label{fig:mnist_max}
\end{figure*}

\begin{figure}[t!]
	\centering
	\begin{subfigure}{0.495\linewidth}
		\includegraphics{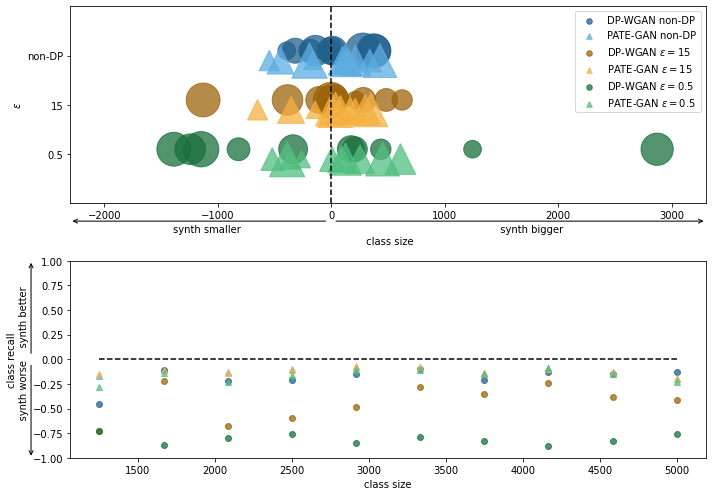}
		\caption{Mixed, Minority (25\%)}
		\label{fig:mnist_mix_min_0.25}
	\end{subfigure}
	\begin{subfigure}{0.495\linewidth}
		\includegraphics{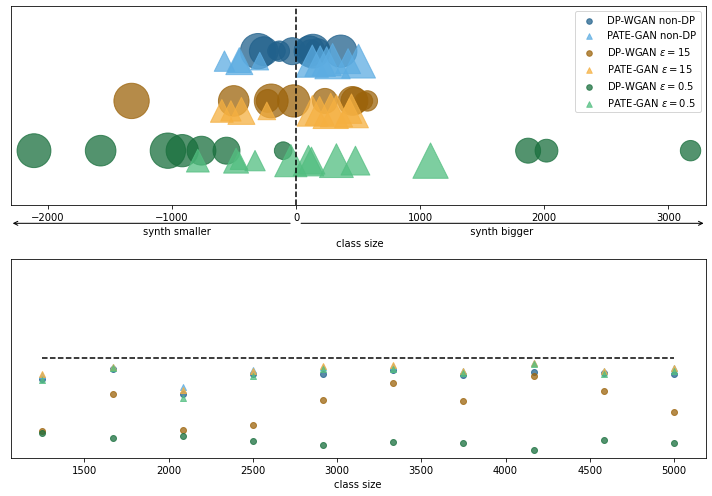}
		\caption{Mixed, Majority (25\%)}
		\label{fig:mnist_mix_max_0.25}
	\end{subfigure}
	\caption{Class size (top) and recall (bottom) of synthetic data, relative to real, generated by DP-WGAN and PATE-GAN trained with different privacy budgets $\epsilon$ on imbalanced MNIST, where ``8'' is minority/majority class accounting to 25\% of the largest/smallest class while all other classes are randomly subsampled in a uniformly decreasing manner.}
	\label{fig:mnist_mix}
\end{figure}

We observe the results in Fig.~\ref{fig:mnist_min} and Tab.~\ref{tab:mnist_min}.
Overall, PATE-GAN exhibits far better performance for both imbalances -- it preserves the counts even for lower $\epsilon$ budgets, the recall drop is not so acute, and its standard deviation is much lower.
DP-WGAN recall looks random for $\epsilon=0.5$, which means that the classifiers failed to learn anything, most likely due to bad quality of the synthetic data.

Looking at the minority class ``8'', however, PATE-GAN fails to generate any digits for imbalance 10\%.
Surprisingly, this phenomenon occurs in the ``non-DP'' case as well.
This could be because the teachers fail to pass samples ``8'' classified as real to the student even though when applied to classification, PATE is more robust under similar imbalance levels~\cite{uniyal2021dp}.

While expectedly DP-WGAN's performance monotonically drops both in terms of size and recall with decreasing~$\epsilon$, PATE-GAN's performance actually increases when DP is applied, e.g.,~$\epsilon=15$ and 5 yield better results than ``non-DP'' in terms of size for both imbalances and in terms of recall for imbalance 10\%.
This is most likely due to the fact that the teacher-discriminators are exposed to different subsets of the real data, and as result, do not learn exactly the same distributions as well as the noise added to their votes, which further enables generalization.

\subsection{Majority Class Results}
\label{ssec:max_results}

The results are in Fig.~\ref{fig:mnist_max} and Tab.~\ref{tab:mnist_max}.
For DP-WGAN, there is a significant drop in recall for all undersampled digits, even for the ``non-DP'' case, unlike PATE-GAN, which again has a very stable behavior.

In terms of size, PATE-GAN generates more imbalanced datasets by producing more ``8s'' and uniformly fewer other digits for all $\epsilon$ budgets but again achieves better results for $\epsilon=15$ than ``non-DP''.
Unlike the minority class setting, no classes ``disappear'' when the imbalance is increased from 25\% to 10\%.
For DP-WGAN, increasing the imbalance leads to much worse performance, which allows us to speculate that PATE-GAN needs smaller training data to capture the underlying distribution and is more robust to imbalance.

Interestingly, some digits suffer a lot more in terms of recall than others (e.g., ``2,'' ``5,'' ``9''), which could be explained because they are visually close to ``8'' but their sizes are much smaller.

\subsection{Mixed Class Results}
\label{ssec:mix_results}

The results are displayed in Fig.~\ref{fig:mnist_mix} and Tab.~\ref{tab:mnist_mix}.
First, we note that with increased privacy PATE-GAN, again, has much lower variability/spread in terms of size and a smaller drop in terms of recall.
We also clearly observe the opposing size effects the two generative models exhibit, similarly to~\cite{ganev2021robin} -- DP-WGAN makes the classes more uniform, i.e., large classes are reduced, and small classes are increased, while PATE-GAN further enforces the imbalance, large classes become even bigger.

In terms of recall, the performance of PATE-GAN for all privacy budgets $\epsilon$ drops slightly in a uniform manner across all classes and actually exceeds the ``non-DP'' DP-WGAN.
This observation hints that underrepresented groups do not suffer unequally under the PATE framework when the imbalance is not severe.
As for DP-WGAN, the size seems to be an important factor as for $\epsilon=15$ smaller classes bear more considerable drop, which is in agreement with previous work~\cite{bagdasaryan2019differential, farrand2020neither}.
Recall for $\epsilon=0.5$ looks random.

\begin{figure*}[t!]
	\centering
	\begin{subfigure}{0.16\linewidth}
		\includegraphics{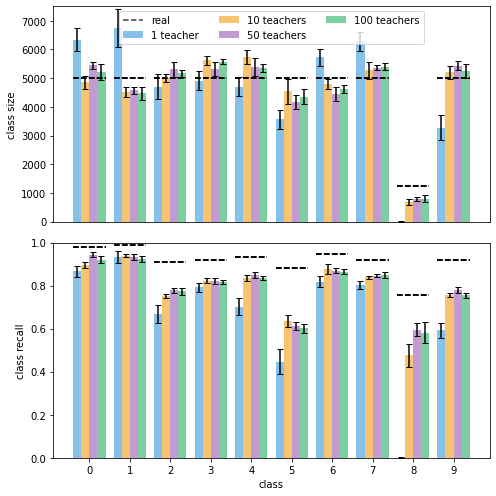}
		\caption{Min (25\%)}
		\label{fig:mnist_teachers_min_0.25}
	\end{subfigure}
	\begin{subfigure}{0.16\linewidth}
		\includegraphics{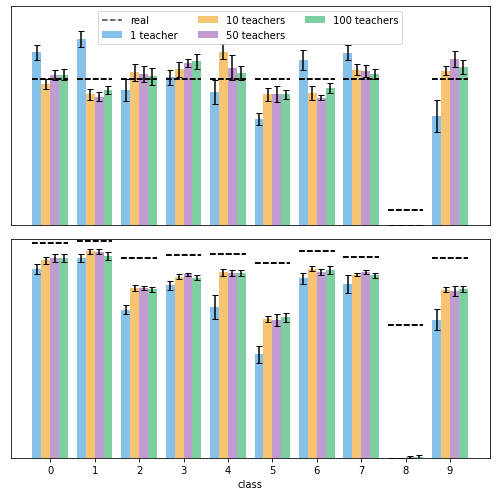}
		\caption{Min (10\%)}
		\label{fig:mnist_teachers_min_0.1}
	\end{subfigure}
	\begin{subfigure}{0.16\linewidth}
		\includegraphics{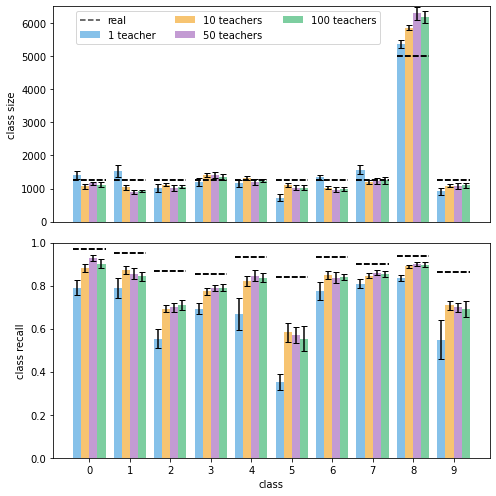}
		\caption{Maj (25\%)}
		\label{fig:mnist_teachers_max_0.25}
	\end{subfigure}
	\begin{subfigure}{0.16\linewidth}
		\includegraphics{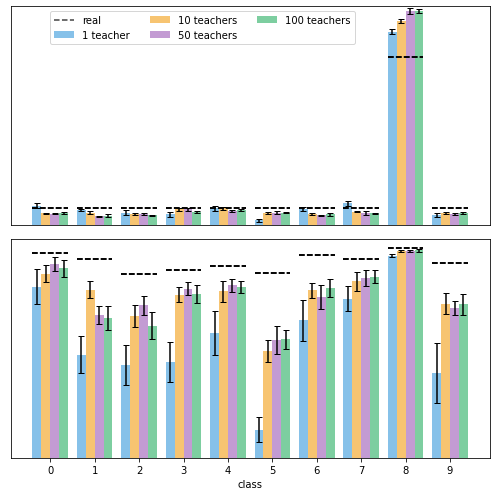}
		\caption{Maj (10\%)}
		\label{fig:mnist_teachers_max_0.1}
	\end{subfigure}
	\begin{subfigure}{0.16\linewidth}
		\includegraphics{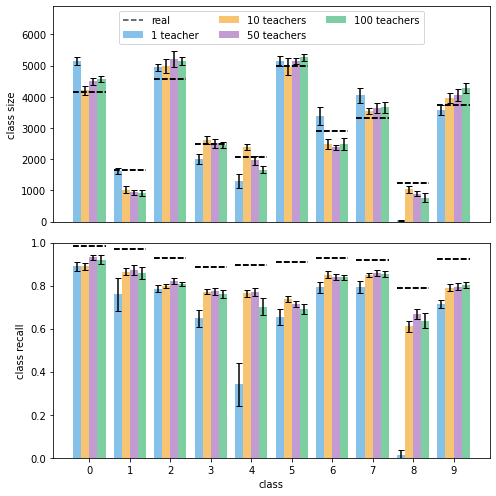}
		\caption{Mix, Min (25\%)}
		\label{fig:mnist_teachers_mix_min_0.25}
	\end{subfigure}
	\begin{subfigure}{0.16\linewidth}
		\includegraphics{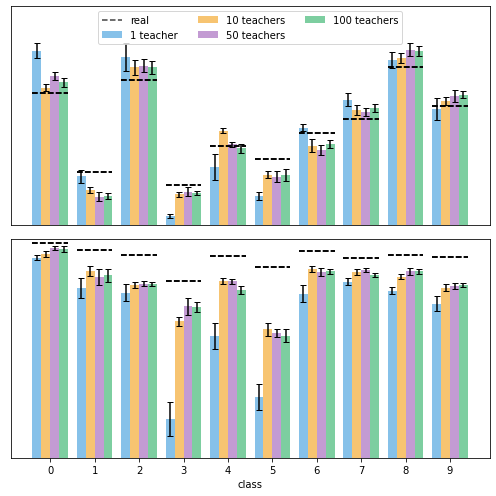}
		\caption{Mix, Maj (25\%)}
		\label{fig:mnist_teachers_mix_max_0.25}
	\end{subfigure}
	\caption{Class size (top) and recall (bottom) of synthetic data generated by PATE-GAN for $\epsilon=5$ in all settings.}
	\label{fig:mnist_teachers}
\end{figure*}

\subsection{Number of Teacher-Discriminators Results}
\label{ssec:teachers_results}

Here, we repeat all the experiments for PATE-GAN with $\epsilon=5$ in the three settings (Minority, Majority, Mixed) but vary the number of teacher-discriminators.
The results can be seen in Fig.~\ref{fig:mnist_teachers} and Tab.~\ref{tab:mnist_teachers}.
Overall, we observe that the best performance is achieved when we set the number of teachers to 10 or 50, with some small exceptions.
This behavior is fully expected and in line with previous work~\cite{jordon2018pate, uniyal2021dp} as having too few/many teachers could fail to generalize or learn at all.
For instance, we see that when we have a single teacher, the standard deviation is much higher across the board.
One interesting case is the Minority setting with imbalance 10\% in Fig.~\ref{fig:mnist_teachers_min_0.1}, where increasing the number of teachers improves performance, and we get the best results for 100.
This still fails to generate any ``8s'', unfortunately.

\section{Related Work}
There is a rich body of research proposing GANs trained with DP guarantees in various domains.
They are predominantly GANs variants utilizing modifications of DP-SGD, e.g., DPGAN~\cite{xie2018differentially} for images and  Electronic Health Records (EHR), dp-GAN~\cite{zhang2018differentially} and DP-CGAN~\cite{torkzadehmahani2019dp} for images, dp-GAN-TSCD~\cite{frigerio2019differentially} for tabular and time series data, etc.
In comparison, there are only a couple of models incorporation PATE into GANs -- PATE-GAN~\cite{jordon2018pate} and G-PATE~\cite{long2021gpate} which ensures DP for the generator by connecting a student generator with an ensemble of teacher discriminators.
Fan~\cite{fan2020survey} provides a more in-depth survey of various DP GANs.

Researchers have studied the disparate effects of DP mechanisms in different contexts.
Kuppam et al.~\cite{kuppam2019fair} and Tran et al.~\cite{tran2021decision} demonstrate that if fund allocations are based on DP statistics, smaller districts could get more resources at the expense of larger ones, compared to what they would receive without DP.

Prior work has also analyzed the disparate effects of DP-SGD applied to deep neural network classifiers~\cite{bagdasaryan2019differential, farrand2020neither, suriyakumar2021chasing} trained on imbalanced datasets.
They empirically demonstrate that the less represented subgroups in the dataset that suffer lower accuracy to start with lose even more utility when DP is applied.
For example, Farrand et al.~\cite{farrand2020neither} show that even small imbalances and loose privacy guarantees could lead to disparate impacts.
Uniyal et al.\cite{uniyal2021dp} find that CNNs trained with PATE suffer from disparate accuracy drops as well, but less severely than with DP-SGD.
There are also papers focusing on learning DP classifiers with fairness constraints~\cite{jagielski2019differentially, tran2021differentially} and on analyzing the PATE framework from a fairness point of view~\cite{tran2021fairness}.
While these efforts consider discriminative models, we examine generative ones.

Finally, analyzing the DP disparity on generative models, Cheng et al.~\cite{cheng2021can} show that training classifiers on balanced DP synthetic images could result in increased majority subgroup influence and utility degradation.
Focusing on tabular data, Pereira et al.~\cite{pereira2021analysis} look at single-attribute subgroup fairness and overall classification while Ganev et al.~\cite{ganev2021robin} analyze class as well as single/multi-attribute subgroup classification parity over a variety of imbalances and privacy budgets.
They find that the disparate effects of DP could be opposing depending on the specific generative model and DP mechanism.

Our work is perhaps closest in spirit to~\cite{uniyal2021dp, ganev2021robin} but we focus on generative models unlike the former and use image data and have a more disciplined approach to constructing the class imbalances, unlike the latter.

\section{Conclusion}
Through our extensive experiments and analysis, we demonstrate that applying DP methods to GANs to preserve the privacy of the data records could lead to disparate effects on class distributions in the generated synthetic data and the performance of downstream tasks.
Overall, PATE exhibits a more desirable behavior than DP-SGD, better privacy-utility trade-off, and while, unfortunately, it still disproportionately affects the minority subparts of the data, it does so to a less severe extend.

\bibliographystyle{abbrv}

%
\appendix

\begin{figure}[h!]
	\centering
	\begin{subfigure}{0.99\linewidth}
		\includegraphics{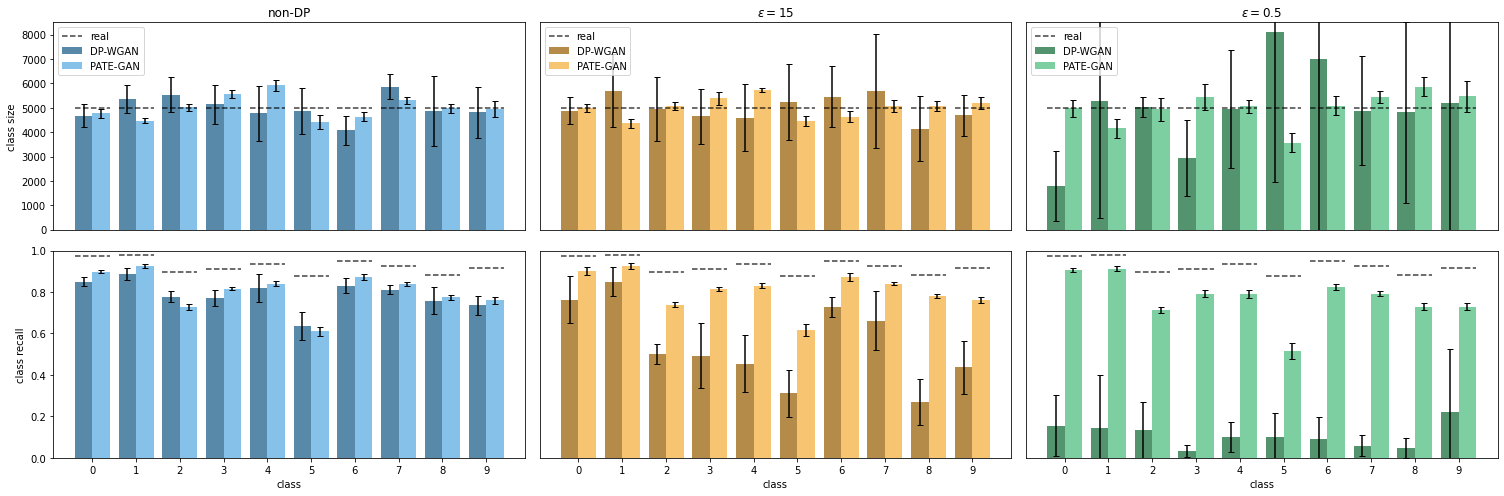}
	\end{subfigure}
	\caption{Class size (top) and recall (bottom) of synthetic data generated by DP-WGAN and PATE-GAN trained with different privacy budgets $\epsilon$ on balanced MNIST.}
	\label{fig:mnist}
\end{figure}

\begin{figure}[h!]
	\centering
	\begin{subfigure}{0.99\linewidth}
		\includegraphics{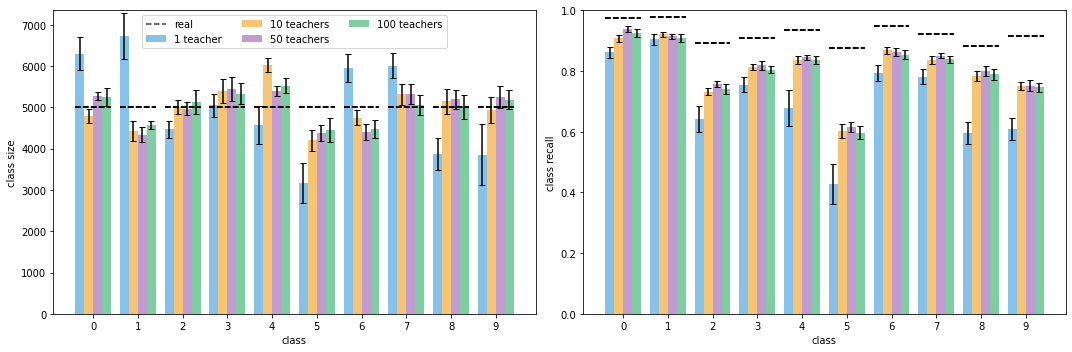}
	\end{subfigure}
	\caption{Class size (left) and recall (right) of synthetic data generated by PATE-GAN for $\epsilon=5$ in balanced settings.}
	\label{fig:mnist_teachers_balanced}
\end{figure}

\section{Balanced Class Results}
\label{sec:balanced}
In this section, we experiment with balanced settings (i.e., we do not artificially imbalance the dataset) to serve as a baseline for the imbalanced settings.
The results are displayed in Fig.~\ref{fig:mnist} and~\ref{fig:mnist_teachers_balanced} as well as Tab.~\ref{tab:mnist} and~\ref{tab:mnist_teachers_balanced}.
In short, yet again, PATE-GAN performs better than DP-WGAN -- it manages to keep the class size balance and utility with increased privacy and has a much lower deviation.
Unlike all other settings, even for the ``non-DP'' case, the size RMSE on synthetic data produced by PATE-GAN is smaller than DP-WGAN.
Interestingly, even with no imbalance, PATE-GAN still benefits from applying a small privacy budget (e.g., $\epsilon=15$).
Similarly to~\ref{ssec:max_results}, for PATE-GAN, the classes that suffer from the biggest accuracy drop with increased privacy are the ones that are visually similar to each other, ``5'', ``8'', ``9'', and ``2''.
These observations allow us to speculate that the issues and effects observed in Sec.~\ref{sec:evaluation} are not only due to the class imbalance but are magnified by it.

\begin{table}[h!]
  \centering
  \begin{tabular}[width=0.99\textwidth]{c | c  c  c  c }
    \hline
    Imbalance   & \multicolumn{4}{c}{Balanced} \\
    $\epsilon$  & no-DP & 15 & 5 & 0.5 \\
    \hline\hline
    Size RMSE  &  & \\
    DP-WGAN & \textbf{473.38} & 487.0 & 1004.19 & 1683.49 \\
    PATE-GAN & 455.78 & \textbf{396.14} & 490.72 & 640.7 \\
    \hline
  Recall TRMSE   &  & \\
    DP-WGAN & \textbf{0.1428} & 0.408 & 0.683 & 0.8173 \\
    PATE-GAN & 0.1316 & \textbf{0.129} & 0.1338 & 0.1729 \\
    \hline
  \end{tabular}
	\caption{Class size RMSE and recall TRMSE summary corresponding to App.~\ref{sec:balanced}, Fig.~\ref{fig:mnist} and ~\ref{fig:mnist_trade_off}.}
	\label{tab:mnist}
\end{table}

\begin{table}[h!]
  \centering
  \begin{tabular}[width=0.99\textwidth]{c | c  }
    \hline
    Imbalance   & Balanced \\
    $\epsilon$  & 5 \\
    \hline\hline
    Size RMSE  & \\
    1 Teacher & 1140.23 \\
    10 Teachers & 490.72 \\
    50 Teachers & 425.23 \\
    100 Teachers & \textbf{354.31} \\
    \hline
  Recall TRMSE   & \\
    1 Teacher & 0.2438 \\
    10 Teachers & 0.1338 \\
    50 Teachers & \textbf{0.1246} \\
    100 Teachers & 0.135 \\
    \hline
  \end{tabular}
	\caption{Class size RMSE and recall TRMSE summary corresponding to App.~\ref{sec:balanced} and Fig.~\ref{fig:mnist_teachers_balanced}.}
	\label{tab:mnist_teachers_balanced}
\end{table}

\section{Tables}
\label{sec:tables}

In this section, we present summary tables of all experiments with imbalanced settings in Sec.~\ref{sec:evaluation}.

\begin{table*}[ht!]
  \centering
  \begin{tabular}[width=0.99\textwidth]{c | c c c c  | c c c c }
    \hline
    Imbalance   & \multicolumn{4}{c}{Minority (25\%)} & \multicolumn{4}{c}{Minority (10\%)} \\
    $\epsilon$  & no-DP & 15 & 5 & 0.5 & no-DP & 15 & 5 & 0.5 \\
    \hline\hline
    Size RMSE  &  &  &  &  &  &  &  & \\
    DP-WGAN & \textbf{433.95} & 835.85 & 1594.41 & 2537.48 & \textbf{303.24} & 681.29 & 921.0 & 1902.04 \\
    PATE-GAN & 456.18 & 433.06 & \textbf{419.82} & 480.61 & 519.16 & 489.0 & \textbf{477.81} & 558.09 \\
    \hline
  Recall TRMSE   &  &  &  &  &  &  &  & \\
    DP-WGAN & \textbf{0.1406} & 0.4038 & 0.6667 & 0.8201 & \textbf{0.1396} & 0.3913 & 0.7205 & 0.8305 \\
    PATE-GAN & \textbf{0.1353 } & \textbf{0.1353} & 0.1363 & 0.1743 & 0.1435 & 0.1412 & \textbf{0.1407} & 0.1746 \\
    \hline
  \end{tabular}
	\caption{Class size RMSE and recall TRMSE summary corresponding to Sec.~\ref{ssec:min_results}, Fig.~\ref{fig:mnist_min} and ~\ref{fig:mnist_trade_off}.}
	\label{tab:mnist_min}
\end{table*}

\begin{table*}[ht!]
  \centering
  \begin{tabular}[width=0.99\textwidth]{c | c c c c  | c c c c }
    \hline
    Imbalance   & \multicolumn{4}{c}{Majority (25\%)} & \multicolumn{4}{c}{Majority (10\%)} \\
    $\epsilon$  & no-DP & 15 & 5 & 0.5 & no-DP & 15 & 5 & 0.5 \\
    \hline\hline
    Size RMSE  &  &  &  &  &  &  &  & \\
    DP-WGAN & 469.23 & \textbf{407.4} & 641.74 & 2038.68 & \textbf{242.38} & 499.73 & 935.04 & 3048.58 \\
    PATE-GAN & 476.98 & 441.95 & 494.13 & \textbf{393.46} & 771.72 & 749.01 & 777.01 & \textbf{477.82} \\
    \hline
  Recall TRMSE   &  &  &  &  &  &  &  & \\
    DP-WGAN & \textbf{0.2961} & 0.5702 & 0.7928 & 0.8291 & \textbf{0.5114} & 0.5853 & 0.7326 & 0.8129 \\
    PATE-GAN & 0.1152 & \textbf{0.1128} & 0.1157 & 0.1529 & 0.1242 & 0.1248 & \textbf{0.1237} & 0.1537 \\
    \hline
  \end{tabular}
	\caption{Class size RMSE and recall TRMSE summary corresponding to Sec.~\ref{ssec:max_results}, Fig.~\ref{fig:mnist_max} and ~\ref{fig:mnist_trade_off}.}
	\label{tab:mnist_max}
\end{table*}

\begin{table*}[ht!]
  \centering
  \begin{tabular}[width=0.99\textwidth]{c | c c c c  | c c c c }
    \hline
    Imbalance   & \multicolumn{4}{c}{Mixed, Minority (25\%)} & \multicolumn{4}{c}{Mixed, Majority (25\%)} \\
    $\epsilon$  & no-DP & 15 & 5 & 0.5 & no-DP & 15 & 5 & 0.5 \\
    \hline\hline
    Size RMSE  &  &  &  &  &  &  &  & \\
    DP-WGAN & \textbf{267.49} & 500.89 & 554.88 & 1383.69 & \textbf{221.31} & 630.52 & 594.04 & 1554.73 \\
    PATE-GAN & 294.14 & \textbf{264.34} & 291.2 & 369.03 & 364.75 & 367.94 & \textbf{348.8} & 552.61 \\
    \hline
  Recall TRMSE   &  &  &  &  &  &  &  & \\
    DP-WGAN & \textbf{0.1809} & 0.4283 & 0.7468 & 0.8138 & \textbf{0.1731} & 0.46 & 0.7093 & 0.8447 \\
    PATE-GAN & 0.1304 & 0.1297 & \textbf{0.1254} & 0.167 & \textbf{0.1264} & 0.1294 & 0.1281 & 0.1674 \\
    \hline
  \end{tabular}
	\caption{Class size RMSE and recall TRMSE summary corresponding to Sec.~\ref{ssec:mix_results}, Fig.~\ref{fig:mnist_mix} and ~\ref{fig:mnist_trade_off}.}
	\label{tab:mnist_mix}
\end{table*}

\begin{table*}[ht!]
  \centering
  \begin{tabular}[width=0.99\textwidth]{c | c  c | c  c | c  c }
    \hline
    Imbalance   & Min (25\%) & Min (10\%) & Maj (25\%) & Maj (10\%) & Mix, Min (25\%) & Mix, Maj (25\%) \\
		$\epsilon$  & 5 & 5 & 5 & 5 & 5 & 5 \\
    \hline\hline
    Size RMSE  &  &  &  &  &  & \\
    1 Teacher & 1165.98 & 929.45 & \textbf{307.98} & \textbf{556.47} & 589.25 & 728.41 \\
    10 Teachers & \textbf{419.82} & 477.81 & 494.13 & 777.01 & \textbf{291.2} & \textbf{348.8} \\
    50 Teachers & 475.89 & 484.62 & 733.51 & 994.75 & 397.74 & 460.87 \\
    100 Teachers & 423.28 & \textbf{364.32} & 677.3 & 988.51 & 440.97 & 410.11 \\
    \hline
  Recall TRMSE  &  &  &  &  &  & \\
    1 Teacher & 0.2576 & 0.2283 & 0.2268 & 0.2862 & 0.2737 & 0.2673 \\
    10 Teachers & 0.1363 & 0.1407 & 0.1157 & 0.1237 & 0.1254 & 0.1281 \\
    50 Teachers & \textbf{0.1269} & 0.1411 & \textbf{0.1135} & \textbf{0.1227} & \textbf{0.1216} & \textbf{0.1241} \\
    100 Teachers & 0.1359 & \textbf{0.1405} & 0.1187 & 0.1281 & 0.137 & 0.13 \\
    \hline
  \end{tabular}
	\caption{Class size RMSE and recall TRMSE summary corresponding to Sec.~\ref{ssec:teachers_results} and Fig.~\ref{fig:mnist_teachers}.}
	\label{tab:mnist_teachers}
\end{table*}

\end{document}